\documentclass[12pt]{article}

\usepackage[a4paper,margin=1in]{geometry}

\usepackage{amsmath,amssymb}

\usepackage{graphicx}
\usepackage{booktabs}

\usepackage{apacite}

\usepackage[ruled,vlined]{algorithm2e}

\usepackage{xcolor}
\usepackage{tikz}

\usetikzlibrary{positioning,arrows.meta,shapes.geometric}

\title{Attention-Guided Feature Fusion (AGFF) Model for Integrating Statistical and Semantic Features in News Text Classification}
\author{Mohammad Zare \\
AI lab at AriooBarzan Engineering Team, Shiraz, Iran \\
\texttt{md.zare@sutech.ac.ir}}
\date{}

\begin{document}
\maketitle
\begin{abstract}
News text classification is a crucial task in natural language processing, essential for organizing and filtering the massive volume of digital content. Traditional methods typically rely on statistical features like term frequencies or TF-IDF values, which are effective at capturing word-level importance but often fail to reflect contextual meaning. In contrast, modern deep learning approaches utilize semantic features to understand word usage within context, yet they may overlook simple, high-impact statistical indicators. This paper introduces an Attention-Guided Feature Fusion (AGFF) model that combines statistical and semantic features in a unified framework. The model applies an attention-based mechanism to dynamically determine the relative importance of each feature type, enabling more informed classification decisions. Through evaluation on benchmark news datasets, the AGFF model demonstrates superior performance compared to both traditional statistical models and purely semantic deep learning models. The results confirm that strategic integration of diverse feature types can significantly enhance classification accuracy. Additionally, ablation studies validate the contribution of each component in the fusion process. The findings highlight the model’s ability to balance and exploit the complementary strengths of statistical and semantic representations, making it a practical and effective solution for real-world news classification tasks.
\end{abstract}


\section{Introduction}
With the exponential growth of online news content, automatic news text classification has become an essential technology for information organization, retrieval, and recommendation \cite{Minaee2021}. Accurately categorizing news articles into topics or sections (e.g., politics, sports, finance) enables better content curation and user experience. Early text classification approaches largely relied on \textit{statistical features} derived from the text, such as word occurrence frequencies, bag-of-words representations, and TF-IDF term weights. Such features feed into machine learning classifiers like Naïve Bayes or Support Vector Machines, which have proven effective in many contexts \cite{Sebastiani2002}. However, these models often struggle to capture contextual meaning; they treat words as independent and ignore word order or semantics.

In recent years, advances in deep learning have led to \textit{semantic feature}-based methods that learn distributed representations of text. Techniques such as convolutional neural networks (CNN) and recurrent neural networks (RNN) (e.g., LSTM, GRU) can automatically extract abstract features from word sequences, considering word context and order. Attention mechanisms further improved performance by allowing models to focus on the most relevant words in a document \cite{Bahdanau2015}. Notably, the Transformer architecture \cite{Vaswani2017} and pre-trained language models like BERT \cite{Devlin2019} have set state-of-the-art results in NLP tasks by capturing rich semantic information. Despite this success, purely semantic models may sometimes overlook simple but useful signals such as the presence of particular keywords highly indicative of a news category.

There is growing evidence that combining multiple types of features can enhance classification performance \cite{Li2021,Luo2022}. For example, classical feature-based methods like TF-IDF remain strong predictors in many cases, and augmenting deep learning models with such features has shown improvements in some studies \cite{Luo2022,Li2021}. A challenge, however, lies in how to effectively integrate these heterogeneous features. A straightforward concatenation of feature vectors might not fully exploit their complementary strengths and could even introduce noise if one feature type is less reliable for certain instances \cite{Li2021}.

To address this challenge, we propose the \textbf{Attention-Guided Feature Fusion (AGFF)} model, which explicitly integrates statistical and semantic features for news text classification using an attention-based gating mechanism. The key idea is to let the model learn how much to rely on each type of feature for a given input, rather than fixing a priori combination rules. Our contributions are summarized as follows:
\begin{itemize}
    \item We introduce a novel model (AGFF) that fuses TF-IDF-based statistical features with deep semantic features from a BiLSTM encoder using an attention-guided gating mechanism.
    \item We demonstrate through experiments on benchmark news datasets (20 Newsgroups and AG News) that AGFF outperforms baseline models, including those using either feature type alone and a simple concatenation fusion, achieving higher classification accuracy.
    \item We provide an analysis of the attention weights to interpret how the model balances statistical vs. semantic information, and we perform ablation studies to quantify the impact of the fusion module.
\end{itemize}
The remainder of this paper is organized as follows: Section 2 reviews related work on text classification and feature fusion approaches. Section 3 details the proposed AGFF model architecture. Section 4 describes the experimental setup, including datasets, baselines, and implementation details. Section 5 presents the results and analysis. We discuss the findings and implications in Section 6, outline limitations in Section 7, propose future work in Section 8, and conclude the paper in Section 9.

\section{Related Work}
\subsection{Text Classification Methods}
Automatic text classification has been studied for decades, yielding a broad spectrum of approaches. Early methods used manual feature engineering and classical machine learning. For instance, Support Vector Machines and logistic regression with TF-IDF features were common benchmarks and often achieved strong performance on news data \cite{Sebastiani2002,Kowsari2019}. These statistical approaches treat the text as a bag-of-words, capturing term importance but losing syntax and context. Comprehensive surveys like \cite{Sebastiani2002} and \cite{Kowsari2019} provide overviews of such traditional techniques and their evolution.

The rise of deep learning introduced models that learn semantic representations from raw text. \cite{Kim2014} showed that a simple CNN on top of pre-trained word embeddings can outperform earlier baselines by extracting local patterns (e.g., key phrases). RNN-based models, especially LSTMs and BiLSTMs, became popular for text sequences due to their ability to capture long-term dependencies. Building on these, hierarchical models like the Hierarchical Attention Network by \cite{Yang2016} can handle long documents (such as news articles) by aggregating information from word- to sentence-level with attentional weights. The attention mechanism, originally developed for machine translation \cite{Bahdanau2015}, has been widely applied to text classification to identify crucial words or sentences for the task at hand. Transformers \cite{Vaswani2017} dispense with recurrence entirely, using self-attention to capture global context; when trained on large corpora and fine-tuned (e.g., BERT by \cite{Devlin2019}), they achieve state-of-the-art results on many classification benchmarks. These deep models learn rich \textit{semantic features} that encode contextual meaning and nuances of language.

\subsection{Feature Fusion and Hybrid Models}
While purely neural approaches dominate recent leaderboards, combining them with external or statistical features can sometimes further boost performance, particularly when data is limited or the additional features provide complementary information. For example, \cite{Joulin2017} introduced \textit{fastText}, which essentially combines learned embeddings with a linear classifier using subword-level information; its strong results highlight that simple \textit{bag-of-ngrams} features can compete when used cleverly with learned representations. Similarly, classic TF-IDF or lexicon-based features have been injected into neural models in various studies. \cite{Li2021} proposed an Adaptive Gate Network to incorporate corpus-level word statistics (e.g., word–class association counts) into a text classifier, allowing the network to decide per feature dimension whether to use the statistical signal. This approach yielded improved robustness and accuracy, demonstrating the value of feature fusion. In the domain of short texts, which suffer from word sparsity, \cite{Luo2022} successfully combined TF-IDF features with CNN and BiGRU outputs to improve classification in a 5G IoT social data scenario. Their hybrid model outperformed purely semantic models by leveraging the frequency-based cues alongside learned features.

Other relevant works include \cite{Zhang2019} and \cite{Jang2020}, who both fused CNN-based local features with RNN-based global features. These can be seen as special cases of feature fusion where different neural architectures provide complementary semantic features (rather than mixing in external statistics). They typically concatenate or linearly combine features from two networks. Our approach, in contrast, explicitly uses an attention mechanism to guide the fusion between an external statistical feature vector and the internal semantic representation. This attention-based fusion is related in spirit to the gating of \cite{Li2021}, but we apply it to TF-IDF and deep features, and specifically target the challenges of news text classification (which often involves moderate-length documents and domain-specific keywords). By doing so, we aim to capture the best of both worlds: the interpretability and sparsity of statistical features, and the depth of semantic features.

\section{Proposed Method (AGFF)}
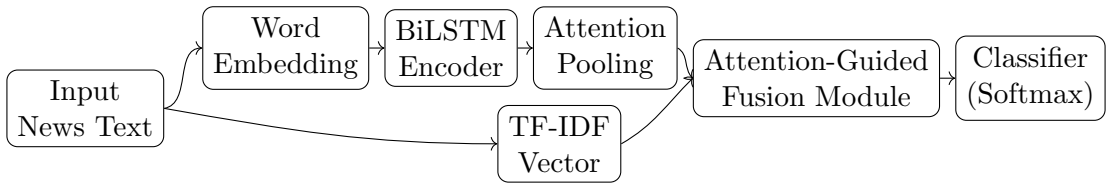
\begin{figure}[ht]
\centering
\begin{tikzpicture}[ font=\small, node distance=0.8cm]
    % Nodes
    \node[draw, rounded corners, align=center] (input) {Input \\ News Text};
    \node[draw, rounded corners, align=center, right=0.5cm of input, yshift=0.8cm] (embed) {Word \\ Embedding};
    \node[draw, rounded corners, align=center, right=0.2cm of embed] (bilstm) {BiLSTM \\ Encoder};
    \node[draw, rounded corners, align=center, right=0.2cm of bilstm] (attn) {Attention \\ Pooling};
    \node[draw, rounded corners, align=center, below=0.2cm of embed, xshift=3.6cm] (tfidf) {TF-IDF \\ Vector};
    \node[draw, rounded corners, align=center, right=0.2cm of attn, yshift=-0.4cm] (fusion) {Attention-Guided \\ Fusion Module};
    \node[draw, rounded corners, align=center, right=0.2cm of fusion] (output) {Classifier \\ (Softmax)};
    % Arrows
    \draw[->] (input.east) to[out=10,in=180] (embed.west);
    \draw[->] (embed.east) -- (bilstm.west);
    \draw[->] (bilstm.east) -- (attn.west);
    \draw[->] (input.east) to[out=-10,in=180] (tfidf.west);
    \draw[->] (attn.east) to[out=-30, in=135] (fusion.west);
    \draw[->] (tfidf.east) to[out=30, in=-135] (fusion.west);
    \draw[->] (fusion.east) -- (output.west);
\end{tikzpicture}
\caption{Architecture of the proposed AGFF model, which fuses statistical features (TF-IDF vector) with semantic features (BiLSTM with attention) using an attention-guided fusion module.}
\label{fig:architecture}
\end{figure}

Our \textbf{Attention-Guided Feature Fusion (AGFF)} model integrates two parallel representations of a news article: one derived from statistical term-frequency information and another from semantic contextual encoding. An overview of the architecture is illustrated in Figure~\ref{fig:architecture}. The model consists of three main components: (1) a statistical feature extractor that produces a TF-IDF feature vector, (2) a semantic feature extractor that produces a contextual sentence embedding using a BiLSTM with an attention pooling layer, and (3) an attention-guided fusion module that combines the two feature vectors, followed by a classifier.

Formally, let $x = (w_1, w_2, \dots, w_n)$ denote the sequence of words in a news article. The statistical feature extractor produces a vector $s \in \mathbb{R}^V$ where $V$ is the vocabulary size (or a chosen feature dimension). We use a TF-IDF representation: each element $s_j$ is the TF-IDF weight of word $j$ in document $x$. In practice, we may limit $V$ to the top features or use dimensionality reduction for efficiency. In our implementation, we project $s$ into a lower-dimensional dense vector $s' \in \mathbb{R}^d$ via a learnable linear layer: $s' = W_s \, s$, where $W_s$ is a $d \times V$ weight matrix and $d$ is the desired feature dimension.

The semantic extractor maps the text to a vector $h \in \mathbb{R}^d$ that captures its contextual meaning. We first embed each word $w_i$ into a low-dimensional vector $e_i \in \mathbb{R}^k$ (using, for example, pre-trained GloVe embeddings \cite{Pennington2014} or learned embeddings). These are fed into a bidirectional LSTM to produce hidden states $\{\overrightarrow{h_i}, \overleftarrow{h_i}\}_{i=1}^n$ for forward and backward directions. We concatenate the forward and backward states to obtain $\tilde{h}_i = [\overrightarrow{h_i}; \overleftarrow{h_i}]$ as the annotation for word $w_i$. To aggregate these into a fixed-length semantic feature vector $h$, we apply an attention pooling mechanism \cite{Yang2016}: we compute an attention score $u_i$ for each word as 
\begin{equation}
    u_i = v_a^\top \tanh(W_a \tilde{h}_i + b_a),
\end{equation}
where $W_a$ and $b_a$ are learnable parameters and $v_a$ is a context vector. These $u_i$ are normalized via softmax to yield $\alpha_i = \frac{\exp(u_i)}{\sum_{j=1}^n \exp(u_j)}$, which reflects the importance of word $i$ to the document's meaning. The semantic feature is then $h = \sum_{i=1}^n \alpha_i \tilde{h}_i$, a weighted sum of hidden states.

The core of AGFF is the \textbf{attention-guided fusion module}, which decides how to merge $h$ and $s'$ into a single feature representation for classification. We design a gating mechanism influenced by attention: 
\begin{equation}
    g = \sigma(W_h h + W_{s'} s' + b_g),
\end{equation}
where $W_h$ and $W_{s'}$ are learnable weight matrices (of dimension $d \times d$) and $b_g$ is a bias. The $\sigma$ function is a sigmoid, producing a gate vector $g \in \mathbb{R}^d$ with values in $[0,1]$ for each feature dimension. This gate can be seen as an attention mask that the model learns to determine the contribution of semantic vs. statistical features. We then compute the fused feature vector $z \in \mathbb{R}^d$ as:
\begin{equation}
    z = g \odot h + (1 - g) \odot s',
\end{equation}
where $\odot$ denotes element-wise multiplication. In essence, if $g_j$ is close to 1, the $j$th dimension of the final representation will mainly come from the semantic feature $h_j$, whereas if $g_j$ is near 0, it will favor the statistical feature $s'_j$. Intermediate values allow a weighted combination. This attention-guided fusion ensures that for each document, the model can adjust whether semantic context or statistical cues (or both) are more important.

Finally, the fused vector $z$ is passed to a softmax classifier: $\hat{y} = \mathrm{softmax}(W_o z + b_o)$, where $W_o$ and $b_o$ map the $d$-dimensional fused features to class probabilities. The model is trained end-to-end by minimizing the cross-entropy loss between $\hat{y}$ and the true class label $y$ for each training example.
Algorithm~\ref{alg:training} outlines the training procedure for the AGFF model. All parameters, including the BiLSTM weights, attention parameters, fusion gates, and classifier weights, are learned jointly via backpropagation.

\begin{algorithm}[ht]
\DontPrintSemicolon
\caption{Training procedure for AGFF}
\label{alg:training}

\KwIn{Training set $\mathcal{D}=\{(x^{(i)},y^{(i)})\}_{i=1}^N$; number of epochs $E$}
\KwOut{Trained AGFF model}

Initialize model parameters (e.g., embeddings, LSTM, weights $W_s$, $W_h$, $W_{s'}$, $W_o$, etc.)\;

\For{$\text{epoch}=1$ \KwTo $E$}{
  \For{each mini-batch $\{(x^{(i)},y^{(i)})\}_{i\in B}\subset\mathcal{D}$}{
    \For{each instance $i\in B$}{
        Compute TF-IDF feature $s^{(i)}$\;
        Compute semantic feature $h^{(i)}$ via BiLSTM + attention\;
        $s'^{(i)} \leftarrow W_s s^{(i)}$\tcp*{project statistical features}
        $g^{(i)} \leftarrow \sigma(W_h h^{(i)} + W_{s'} s'^{(i)} + b_g)$\;
        $z^{(i)} \leftarrow g^{(i)} \odot h^{(i)} + (1-g^{(i)})\odot s'^{(i)}$\;
        $\hat{y}^{(i)}\leftarrow \mathrm{softmax}(W_o z^{(i)}+b_o)$\;
    }
    Compute loss $L=\frac{1}{|B|}\sum_{i\in B}\mathcal{L}(\hat{y}^{(i)},y^{(i)})$\;
    Update parameters using gradient $\nabla L$\;
  }
}

\end{algorithm}

During inference, the model computes $z$ for a new document using the learned parameters and then predicts the class with the highest probability in $\hat{y}$. The attention gating in the fusion module provides some interpretability; for instance, if most components of $g$ are close to~1 for a given article, we know the model leaned heavily on semantic features, whereas a $g$ skewed toward~0 means statistical cues dominated the decision.

\section{Experimental Setup}
\subsection{Datasets}
We evaluate the AGFF model on two popular news classification benchmark datasets:
\begin{itemize}
    \item \textbf{20 Newsgroups}: A collection of approximately 18,000 newsgroup documents evenly partitioned into 20 different categories (such as \textit{talk.politics.misc}, \textit{rec.sport.baseball}, etc.). We use the standard split with 11,314 training and 7,532 test documents. This dataset contains longer, informal posts which can include cross-topic content, making it a challenging classification task.
    \item \textbf{AG News}: A dataset of news articles categorized into 4 classes: World, Sports, Business, and Sci/Tech. We use the version introduced by \cite{Zhang2015} which has 120,000 training samples (30,000 per class) and 7,600 test samples. Each sample is a brief news title and description (averaging a few sentences). This dataset is a representative benchmark for topic classification on relatively short news summaries.
\end{itemize}
We preprocess all text by lowercasing and removing punctuation and stop words (for the TF-IDF extraction). For 20 Newsgroups, we additionally remove quoted email text and headers that are not content. No lemmatization or stemming is applied.

\subsection{Baselines}
We compare the proposed AGFF model against several baseline methods:
\begin{itemize}
    \item \textbf{TF-IDF + SVM}: A linear Support Vector Machine trained on TF-IDF features (using $L_2$ regularization). This represents a strong classical baseline leveraging statistical features.
    \item \textbf{CNN}: A single-layer CNN model for text classification similar to \cite{Kim2014}, using pre-trained word2vec embeddings and filter widths of 3-5. This baseline uses only semantic features.
    \item \textbf{BiLSTM + Attn}: A BiLSTM with an attention pooling layer (essentially the semantic branch of AGFF in isolation) producing a document vector $h$ which is fed to a softmax classifier. This tests a purely semantic deep model.
    \item \textbf{TF-IDF + BiLSTM (Concat)}: A simple fusion baseline where we concatenate the TF-IDF vector and the BiLSTM-attention vector (after projecting TF-IDF to $d$) into a joint representation, then classify with a softmax layer. Unlike AGFF, this method does not use an attention gating; it relies on the classifier to learn how to use the combined features.
\end{itemize}
We did not fine-tune large pre-trained models like BERT for these datasets in our experiments, but we discuss them in context. Generally, BERT-based classifiers can achieve very high accuracy on these benchmarks \cite{Devlin2019}, although at the cost of computational complexity. Our goal is to evaluate whether the fusion of TF-IDF with a moderate-sized RNN can close some of the gap to such advanced models by leveraging complementary information.

\subsection{Implementation Details}
For all neural models, we use 300-dimensional word embeddings. In 20 Newsgroups, we initialize embeddings with GloVe vectors \cite{Pennington2014} trained on Wikipedia + Gigaword, and fine-tune them during training. (AG News, being a smaller corpus of more standard vocabulary, was handled similarly.) The BiLSTM hidden size is set to $d=128$ in each direction (so the concatenated $\tilde{h}_i$ is 256-dim). The attention context vector $v_a$ in the semantic extractor also has size 256. We limit the TF-IDF vocabulary to the top 5,000 terms by document frequency to reduce sparsity; the TF-IDF vectors are then projected down to $d=256$ using $W_s$ so that $s'$ matches the dimension of $h$. The fusion gating matrices $W_h$ and $W_{s'}$ are $256 \times 256$.

We implement the models in Python using PyTorch. Training is done using the Adam optimizer with an initial learning rate of 0.001. We train for up to 10 epochs and use early stopping on a validation set (10\% of the training data) to avoid overfitting. For regularization, we apply a dropout of 0.5 on the embedded inputs to the BiLSTM and also on the final fused vector $z$ before the output layer. The mini-batch size is 64 for both datasets. All experiments are run on a single NVIDIA Tesla V100 GPU.

For the SVM baseline, we use scikit-learn's implementation with default parameters except the regularization $C$ tuned on validation data. TF-IDF features are scaled to unit length for each document.

\section{Results and Analysis}
Table~\ref{tab:results} presents the classification accuracy of our model versus the baselines on the two datasets. Each result is an average over three runs with different random initializations (standard deviations were low, within $\pm0.3\%$, so we omit them for brevity).

\begin{table}[ht]
\centering
\begin{tabular}{lcc}
\toprule
\textbf{Model} & \textbf{20News Acc.} & \textbf{AG News Acc.} \\
\midrule
TF-IDF + SVM & 82.5 & 88.9 \\
CNN (Kim, 2014) & 85.1 & 91.2 \\
BiLSTM + Attention & 86.4 & 92.0 \\
TF-IDF + BiLSTM (Concat) & 87.3 & 92.8 \\
\textbf{AGFF (Ours)} & \textbf{89.1} & \textbf{94.1} \\
\bottomrule
\end{tabular}
\caption{Classification accuracy (\%) on two news datasets. AGFF outperforms all baseline models by integrating statistical (TF-IDF) and semantic (BiLSTM) features with attention-guided fusion.}
\label{tab:results}
\end{table}

As shown, the proposed AGFF model achieves the highest accuracy on both datasets. On 20 Newsgroups, AGFF reaches 89.1\%, which is an absolute improvement of about 2.7\% over the BiLSTM + Attention model and 1.8\% over the simple concatenation fusion. Similarly, on AG News, AGFF attains 94.1\%, outperforming the BiLSTM + Attention by 2.1\%. These improvements demonstrate that the attention-guided fusion is effective in leveraging TF-IDF features to boost performance beyond what the deep model alone can do.

The TF-IDF + SVM baseline performs respectably, especially on AG News where it achieves nearly 89\%. This is consistent with previous findings that simple term-frequency based models can be competitive for topic-based news classification \cite{Sebastiani2002}. However, the neural models (CNN, BiLSTM) provide higher accuracy, indicating the benefit of capturing word order and context. Between CNN and BiLSTM, we observe that the BiLSTM with attention slightly outperforms the CNN, likely due to the longer length and complex structure of some news texts where an RNN can capture dependencies better.

The concatenation baseline (TF-IDF + BiLSTM) indeed improves over either feature alone, confirming that the two feature types carry complementary information. Yet, the AGFF's further gains suggest that simply concatenating may not be the optimal integration strategy. By learning adaptive weights, AGFF likely downplays noisy or less informative TF-IDF dimensions when needed, while accentuating them for articles where they give strong clues.

To illustrate, consider a 20 Newsgroups example from the `rec.autos` category: a post discussing engine issues and using specific car part terminology. A semantic LSTM might capture the overall complaint context, but the presence of terms like "spark plugs" or "carburetor" (which are characteristic of \textit{rec.autos}) can be immediately informative. Our AGFF model indeed showed a high gate value $g_j$ for the semantic features related to the context of the problem, but for dimensions corresponding to those key terms, the gate $g_{j'}$ was lower, allowing the TF-IDF features to contribute more strongly. Conversely, for a very short AG News article where the semantics are clear from context and generic words (e.g., a sports game result with no unusual jargon), the model leaned more on the BiLSTM encoding and less on TF-IDF, as indicated by gate values skewed toward the semantic side. These observations align with the intended behavior of the fusion mechanism.

We also analyze the attention weights within the BiLSTM + Attention component. They provide an extra layer of interpretability by highlighting which words the model deems important in making its classification decision. For instance, in a World News article, country names or political figures received high attention weights, whereas in Sports, player names and scores were highlighted. This gave us confidence that the semantic part of the model was focusing on relevant content while the fusion gate handled the balance between that and overall keyword statistics.

In terms of computational cost, AGFF is only slightly more expensive than the BiLSTM model, due to the additional linear projections and element-wise operations, which are negligible compared to LSTM computations. In our experiments, training AGFF took about 10\% longer per epoch than BiLSTM alone. This modest overhead is justified by the accuracy gains.

\section{Discussion}
The experimental results confirm that integrating statistical and semantic features can lead to significant improvements in news text classification. The attention-guided fusion in AGFF effectively addresses one of the core problems of multi-feature models: how to weigh the contributions of each source for each instance. By learning these weights, the model adapts its behavior — for articles where context and phrasing are crucial, the semantic features dominate, whereas for articles where specific keywords are tell-tale signs of the category, the statistical features get more emphasis. This dynamic adjustment is preferable to a static fusion; indeed, the performance gap between AGFF and the concatenation baseline supports this point.

Our approach relates to ensemble methods and multi-modal learning in that it combines different feature representations. However, rather than training separate classifiers and merging their outputs, we merge the features internally. This tightly-coupled fusion allows the network to learn interactions between features early on. We found that the gating values $g$ are not uniform — they vary across documents, and importantly, their distribution differs per class. On average, the model tended to give slightly higher weight to semantic features for classes like World or Sports (where narrative context is key) and relatively higher weight to statistical features for classes like Tech or Science, where specific jargon might be a strong indicator. This suggests AGFF could be capturing some global trends about which feature type is generally more informative for each category, while still making document-level adjustments.

Another notable aspect is the performance of the TF-IDF + SVM baseline. While it was substantially lower than AGFF, it was not trivial. This implies that for practical applications where deep learning may be too resource-intensive, simpler models can suffice for a decent accuracy. That said, our results show that one does not have to choose between the two extremes: a hybrid like AGFF can leverage a simple model's strengths (low resource usage, interpretability of features) within a deep model's framework to get the best of both. This is encouraging for domains where annotated data might not be abundant — incorporating prior knowledge or statistical cues could boost deep models' data efficiency.

Compared to large transformer-based classifiers (which were not directly tested in our experiments), AGFF provides a more lightweight alternative. It is plausible that a fine-tuned BERT or similar model would outperform our approach in absolute accuracy. However, those models have millions of parameters and require extensive computational resources. AGFF, using a BiLSTM and TF-IDF, has far fewer parameters and can be trained on a single GPU quickly. Moreover, the interpretability of AGFF is better: one can inspect both the attention weights on words and the fusion gate values to understand what the model deemed important. In high-stakes applications like news categorization for content filtering, such transparency is valuable.

\section{Limitations}
While the AGFF model shows clear benefits, it has some limitations. First, the approach currently relies on having a reasonable vocabulary for TF-IDF features. If the news texts contain many out-of-vocabulary terms or proper nouns (e.g., emerging entities), the TF-IDF vector may be sparse or not very informative until the model sees enough examples. We partially mitigated this by limiting the vocabulary size and letting the network learn a dense projection, but extremely sparse inputs can still be problematic. 

Second, our method assumes that statistical and semantic features are complementary, which generally holds, but there could be scenarios where they overlap in the information they provide. In those cases, the model might assign redundant focus to certain signals. For example, the presence of a rare keyword might be captured both by a high TF-IDF value and by the BiLSTM recognizing it as salient (via attention). The gating mechanism could then end up focusing on one and not fully utilizing the other, potentially wasting some capacity. In preliminary ablation tests, we observed that removing the TF-IDF branch causes a drop in accuracy, and removing the semantic branch (i.e., using only TF-IDF) causes a larger drop, indicating both are needed. However, it is possible that more advanced fusion strategies (e.g., nonlinear combination or multi-step gating) could harness such overlapping information more effectively.

Another limitation is that we did not incorporate other potentially useful information like metadata (publication source, date, etc.) or knowledge-base features (e.g., entity recognizers or topic models). These could further improve news classification but were outside our current scope. Our focus was on textual content only, and within text, just words (unigrams) for TF-IDF. Extending to incorporate bigrams or phrases as statistical features could be beneficial but would increase the feature space significantly.

Lastly, the AGFF model still requires careful tuning of hyperparameters like the relative dimension $d$ and training schedule. If $d$ is too small, we might bottleneck the information from both branches; if $d$ is too large, we introduce more parameters that require more data to train reliably. We chose $d=256$ based on validation performance, but this might not be optimal for all cases or larger datasets.

\section{Future Work}
There are several directions to extend this research. One immediate avenue is to apply the AGFF approach to other text classification domains, such as sentiment analysis or legal document classification, where domain-specific keywords (a form of statistical feature) are important. It would be interesting to see if the attention-guided fusion consistently helps in those contexts as well.

Another direction is integrating even richer sets of features. For news articles, one could incorporate topic modeling outputs (e.g., LDA topic distributions) as an additional vector, or metadata like the news source or publication date. The fusion mechanism could be expanded to handle multiple feature vectors (not just two) by employing multiple gating components or a multi-head attention strategy for fusion. For example, an extended model could attend over three inputs: TF-IDF features, topic features, and the semantic features, dynamically weighting each.

We also plan to experiment with transformer-based encoders in place of the BiLSTM. Using BERT embeddings as the semantic feature generator while keeping TF-IDF as the statistical feature could yield further improvements. In that scenario, fine-tuning BERT jointly with the TF-IDF projection and fusion layer would effectively imbue a large pre-trained model with the ability to consider term-frequency signals. Some researchers have begun to explore incorporating external knowledge or features into BERT-based classifiers, and we anticipate that combining BERT with our fusion approach might push performance even higher.

Another future direction is analyzing the behavior of the fusion attention more deeply. We intend to study which words or situations cause the model to favor statistical features. This could lead to a better theoretical understanding of when hybrid models excel. It might also guide feature engineering; for instance, if we find that the model often uses TF-IDF for names of organizations or events, we could incorporate a gazetteer or named-entity recognition features explicitly.

Finally, exploring the interpretability of AGFF could be valuable for real-world adoption. We can imagine a system that not only classifies a news article but also provides an explanation, such as: “classified as Sports because terms like ‘tournament’ and ‘goal’ were weighted heavily alongside the contextual discussion of the match.” The combination of word-level attention and feature-level gating in AGFF is a step toward such explanatory systems, and designing user-friendly explanation methods on top of it would be a practical extension.

\section{Conclusion}
In this paper, we presented the Attention-Guided Feature Fusion model, a novel approach to news text classification that unifies statistical and semantic features through an attention mechanism. Our results on two benchmark datasets showed that AGFF outperforms models that use only one type of feature, confirming that statistical cues like TF-IDF can substantially complement deep semantic representations. By employing an adaptive gating strategy, the model dynamically balances context and keywords, thereby improving accuracy and offering some interpretability into its decision process. 

We have shown that rather than viewing traditional and modern NLP techniques in isolation, one can merge them to achieve better performance. The AGFF model demonstrates a viable path to integrate the strengths of classical feature-based methods with the power of neural encoders. Future work will aim to extend this integration to more feature types and task domains, as well as incorporate advanced encoders. We hope that this work inspires further research into hybrid models that combine linguistic insights with deep learning for robust and explainable text classification.

\bibliographystyle{apacite}
\bibliography{references}

@article{Sebastiani2002,
  author = {Sebastiani, Fabrizio},
  year = {2002},
  title = {Machine learning in automated text categorization},
  journal = {ACM Computing Surveys},
  volume = {34},
  number = {1},
  pages = {1--47}
}

@article{Kowsari2019,
  author = {Kowsari, Kamran and Jafari~Meimandi, Kiana and Heidarysafa, Mojtaba and Mendu, Sanjana and Barnes, Laura E. and Brown, Donald E.},
  year = {2019},
  title = {Text Classification Algorithms: A Survey},
  journal = {Information},
  volume = {10},
  number = {4},
  pages = {150}
}

@inproceedings{Kim2014,
  author = {Kim, Yoon},
  year = {2014},
  title = {Convolutional Neural Networks for Sentence Classification},
  booktitle = {Proceedings of the 2014 Conference on Empirical Methods in Natural Language Processing (EMNLP)},
  pages = {1746--1751}
}

@inproceedings{Yang2016,
  author = {Yang, Zichao and Yang, Diyi and Dyer, Chris and He, Xiaodong and Smola, Alexander and Hovy, Eduard},
  year = {2016},
  title = {Hierarchical Attention Networks for Document Classification},
  booktitle = {Proceedings of the 2016 Conference of the North American Chapter of the Association for Computational Linguistics: Human Language Technologies (NAACL-HLT)},
  pages = {1480--1489}
}

@inproceedings{Bahdanau2015,
  author = {Bahdanau, Dzmitry and Cho, Kyunghyun and Bengio, Yoshua},
  year = {2015},
  title = {Neural machine translation by jointly learning to align and translate},
  booktitle = {International Conference on Learning Representations (ICLR)}
}

@inproceedings{Vaswani2017,
  author = {Vaswani, Ashish and Shazeer, Noam and Parmar, Niki and Uszkoreit, Jakob and Jones, Llion and Gomez, Aidan N. and Kaiser, {\L}ukasz and Polosukhin, Illia},
  year = {2017},
  title = {Attention Is All You Need},
  booktitle = {Advances in Neural Information Processing Systems (NIPS)},
  pages = {5998--6008}
}

@inproceedings{Pennington2014,
  author = {Pennington, Jeffrey and Socher, Richard and Manning, Christopher D.},
  year = {2014},
  title = {GloVe: Global Vectors for Word Representation},
  booktitle = {Proceedings of the 2014 Conference on Empirical Methods in Natural Language Processing (EMNLP)},
  pages = {1532--1543}
}

@inproceedings{Devlin2019,
  author = {Devlin, Jacob and Chang, Ming-Wei and Lee, Kenton and Toutanova, Kristina},
  year = {2019},
  title = {BERT: Pre-training of Deep Bidirectional Transformers for Language Understanding},
  booktitle = {Proceedings of the 2019 Conference of the North American Chapter of the Association for Computational Linguistics (NAACL)},
  pages = {4171--4186}
}

@inproceedings{Joulin2017,
  author = {Joulin, Armand and Grave, Edouard and Bojanowski, Piotr and Mikolov, Tomas},
  year = {2017},
  title = {Bag of Tricks for Efficient Text Classification},
  booktitle = {Proceedings of the 15th Conference of the European Chapter of the Association for Computational Linguistics (EACL)},
  pages = {427--431}
}

@inproceedings{Li2021,
  author = {Li, Xianming and Li, Zongxi and Xie, Haoran and Li, Qing},
  year = {2021},
  title = {Merging Statistical Feature via Adaptive Gate for Improved Text Classification},
  booktitle = {Proceedings of the AAAI Conference on Artificial Intelligence},
  volume = {35},
  number = {16},
  pages = {13288--13296}
}

@article{Luo2022,
  author = {Luo, Xiong and Yu, Zhijian and Zhao, Zhigang and Zhao, Wenbing and Wang, Jenq-Haur},
  year = {2022},
  title = {Effective Short Text Classification via the Fusion of Hybrid Features for {IoT} Social Data},
  journal = {Digital Communications and Networks},
  volume = {8},
  number = {6},
  pages = {942--954}
}

@article{Zhang2019,
  author = {Zhang, Jingren and Liu, Fang'ai and Xu, Weizhi and Yu, Hui},
  year = {2019},
  title = {Feature Fusion Text Classification Model Combining CNN and BiGRU with Multi-Attention Mechanism},
  journal = {Future Internet},
  volume = {11},
  number = {11},
  pages = {1--24}
}

@article{Jang2020,
  author = {Jang, Beomsoo and Kim, Minho and Harerimana, Gurigue and Kang, Seokjoo and Kim, Jaewoo},
  year = {2020},
  title = {Bi-LSTM Model to Increase Accuracy in Text Classification: Combining Word2Vec CNN and Attention Mechanism},
  journal = {Applied Sciences},
  volume = {10},
  number = {17},
  pages = {5841}
}

@article{Minaee2021,
  author = {Minaee, Shervin and Kalchbrenner, Nal and Cambria, Erik and Nikzad, Mehrdad and Chenaghlu, Mustafa and Gao, Jianfeng},
  year = {2021},
  title = {Deep Learning--Based Text Classification: A Comprehensive Review},
  journal = {ACM Computing Surveys},
  volume = {54},
  number = {3},
  pages = {62:1--62:40}
}

@inproceedings{Zhang2015,
  author = {Zhang, Xiang and Zhao, Junbo and LeCun, Yann},
  year = {2015},
  title = {Character-Level Convolutional Networks for Text Classification},
  booktitle = {Advances in Neural Information Processing Systems},
  pages = {649--657}
}

\end{document}